\documentclass[runningheads]{llncs}
\usepackage{amsmath,amssymb,latexsym}
\usepackage{graphicx}
\newcommand{\q}{\ensuremath{\mathbf{q}}}
\newcommand{\p}{\ensuremath{\mathbf{p}}}
\newcommand{\QQ}{\ensuremath{\mathbf{Q}}}
\newcommand{\PP}{\ensuremath{\mathbf{P}}}
\newcommand{\btheta}{\ensuremath{\boldsymbol{\theta}}}
\newcommand{\bbeta}{\ensuremath{\boldsymbol{\beta}}}

\begin{document}
\title{Learning and Interpreting Potentials for Classical Hamiltonian Systems}
%
%
\author{Harish~S. Bhat \orcidID{0000-0001-7631-1831} \thanks{This work was performed under the auspices of the U.S. Department of Energy by Lawrence Livermore National Laboratory under Contract DE-AC52-07NA27344 and was supported by the LLNL-LDRD Program under Project No. 19-ERD-009. LLNL-PROC-779792.}}
\authorrunning{H.~S. Bhat}
%
\institute{Applied Mathematics Unit, University of California, Merced, USA
\email{hbhat@ucmerced.edu}}

\maketitle              
\begin{abstract}
We consider the problem of learning an interpretable potential energy function from a Hamiltonian system's trajectories.  We address this problem for classical, separable Hamiltonian systems.  Our approach first constructs a neural network model of the potential and then applies an equation discovery technique to extract from the neural potential a closed-form algebraic expression.  We demonstrate this approach for several systems, including oscillators, a central force problem, and a problem of two charged particles in a classical Coulomb potential.    Through these test problems, we show close agreement between learned neural potentials, the interpreted potentials we obtain after training, and the ground truth.  In particular, for the central force problem, we show that our approach learns the correct effective potential, a reduced-order model of the system.

\end{abstract}
\section{Introduction}
As a cornerstone of classical physics, Hamiltonian systems arise in numerous settings in engineering and the physical sciences.  Common examples include coupled oscillators, systems of particles/masses subject to classical electrostatic or gravitational forces, and rigid bodies.   For integer $d \geq 1$, let $\q(t) \in \mathbb{R}^d$ and $\p(t) \in \mathbb{R}^d$ denote, respectively, the position and momentum of the system at time $t$.  Let $T$ and $V$ denote kinetic and potential energy, respectively.  Our focus here is on classical, separable systems that arise from the Hamiltonian
\begin{equation}
\label{eqn:H}
H(\p, \q) =T(\p) + V(\q).
\end{equation}
\emph{In this paper, we consider the problem of learning or identifying the potential energy $V$ from data $( \q(t), \p(t) )$ measured at a discrete set of times.}  We assume $T$ is known.  To motivate this problem, consider the setting of $m$ interacting particles in three-dimensional space; here $d = 3m$.  Suppose that we are truly interested in a reduced set of variables, e.g., the position and momentum of one of the $m$ particles.  Let us denote the reduced-order quantities of interest by $(\widetilde{\q}(t), \widetilde{\p}(t)) \in \mathbb{R}^{2 \widetilde{d}}$.   The direct approach is to integrate numerically the $6m$-dimensional system of differential equations for the full Hamiltonian (\ref{eqn:H}) and then use the full solution $( \q(t), \p(t) )$ to compute  $(\widetilde{\q}(t), \widetilde{\p}(t))$.  While such an approach yields numerical answers, typically, \emph{it does not explain} how the reduced-order system evolves dynamically in time.  If we suspect that $(\widetilde{\q}(t), \widetilde{\p}(t))$ itself satisfies a Hamiltonian system, we can search for a potential $\widetilde{V}(\widetilde{\q})$ that yields an accurate, reduced-order model for $(\widetilde{\q}(t), \widetilde{\p}(t))$. If $\widetilde{V}$ is interpretable, we can use it to explain the reduced system's dynamics---\emph{here we mean interpretability in the sense of traditional models in the physical sciences, which are written as algebraic expressions}, not as numerical algorithms.  We can also use the reduced-order model to simulate $(\widetilde{\q}(t), \widetilde{\p}(t))$ directly, with computational savings that depend on $d/\widetilde{d}$.

There is a rapidly growing literature on machine learning of potential energies in computational/physical chemistry, e.g., \cite{Behler07,Hansen15,Ramakrishnan15,Behler2016,Artrith2016}.  As in these studies, the present work uses neural networks to parameterize the unknown potential.  A key difference is that, in the present work, we apply additional methods to interpret the learned neural potential.  There exists a burgeoning, recent literature on learning interpretable dynamical systems from time series, e.g., \cite{Brunton2016,bhat_nonparametric_2016,BhatDale2018,Sahoo2018,Duncker2019}, to cite but a few.  We repurpose one such method---SINDy (sparse identification of nonlinear dynamics)---to convert the learned neural potential into a closed-form algebraic expression that is as interpretable as classical models.  We apply only one such method for accomplishing this conversion into an algebraic expression; we hope that the results described here lead to further investigation in this area.

\section{Approach}
Assume $T(\p) = \sum_{i=1}^d M_{ii}^{-1} p_i^2$ where $M$ is a diagonal mass matrix.  Then, from (\ref{eqn:H}), we can write Hamilton's equations:
\begin{subequations}
\label{eqn:ham}
\begin{align}
\label{eqn:qdot}
\dot{\q} &= M^{-1} \p \\
\label{eqn:pdot}
\dot{\p} &= -\nabla V(\q).
\end{align}
\end{subequations}
Let the training data consist of a set of $R$ trajectories; the $j$-th such trajectory is $\{ \q^j_i, \p^j_i \}_{i=0}^{N}$.  Here $(\q^j_i, \p^j_i)$   denotes a measurement of $(\q(t), \p(t))$ at time $t = i h$ for fixed $h > 0$.  We choose this equispaced temporal grid for simplicity; this choice is not essential.    Because we treat the kinetic energy $T$ as known, we assume that the training data consists of (possibly noisy) measurements of a system that satisfies (\ref{eqn:qdot}).  We now posit a model for $V$ that depends on a set of parameters $\btheta$.  For instance, if we model $V$ using a neural network, $\btheta$ stands for the collection of all network weights and biases.  Then we use (\ref{eqn:pdot}) to  form an empirical risk loss 
\begin{equation}
\label{eqn:risk}
L(\btheta) = \frac{1}{R N} \sum_{j=1}^{R} \sum_{i=0}^{N-1} \left \| \frac{\p^j_{i+1} - \p^j_i}{h} + \nabla_{\q} V(\q^j_i; \btheta) \right \|^2.
\end{equation}
Let $\tau = Nh$ denote the final time in our grid.  Note that (\ref{eqn:risk}) approximates $E \left[ (1/\tau) \int_{t=0}^{t=\tau} \| \dot{\PP}(t) + \nabla_{\QQ} V(\QQ(t); \btheta) \|^2 \mathrm{dt} \right]$, the expected mean-squared error of a random trajectory $(\QQ(t),\PP(t))$ assumed to satisfy (\ref{eqn:qdot}).

We model $V$ using a dense, feedforward neural network with $L \geq 2$ layers.  Because we train with multiple trajectories, the input layer takes data in the form of two tensors---one for $\q$ and one for $\p$---with dimensions $N \times R \times d$.  The network then transposes and flattens the data to be of dimension $N R \times d$.  Thus begins the potential energy function part of the network (referred to in what follows as the \emph{neural potential}), which takes a $d$-dimensional vector as input and produces a scalar as output.  Between the neural potential's $d$-unit input layer and $1$-unit output layer, we have a number of hidden layers.  In this model, we typically choose hidden layers to all have $\nu$ units where $1 \leq \nu \leq d$.  As these architectural details differ by example, we give them below.

Note that the loss (\ref{eqn:risk}) involves the gradient of $V$ with respect to the input $\mathbf{q}$.  We use automatic differentiation to compute this gradient.  More specifically, in our IPython/Jupyter notebooks (linked below), we use the \texttt{batch\_jacobian} method in TensorFlow.  This  is easy to implement, fast, and accurate up to machine precision.

The trained network gives us a neural potential $\widehat{V} : \mathbb{R}^d \to \mathbb{R}$.  To interpret $\widehat{V}$, we apply the SINDy method \cite{Brunton2016}.  We now offer a capsule summary of this technique.  Suppose we have a grid $\{\mathbf{x}^k\}_{k=1}^K$ of points in $\mathbb{R}^d$.  We use the notation $\mathbf{x}^k = (x_1^k, \ldots, x_d^k)$.  We evaluate $\widehat{V}$ on the grid, resulting in a vector of values that we denote by $\mathbf{V}$.  We also evaluate on the grid a library of $J$ candidate functions $\xi^j : \mathbb{R}^d \to \mathbb{R}$ for $1 \leq j \leq J$; each such evaluation results in a vector $\Xi^j$ that we take to be the $j$-th column of a matrix $\Xi$.  In $d=1$, examples of candidate functions are $\{1, x, x^2, \ldots\}$ or $\{1, x^{-1}, x^{-2}, \ldots\}$.  In $d=2$, an example is $\{1, x_1, x_2, x_1^2, x_1 x_2, x_2^2, \ldots\}$.  Each candidate function is simply a scalar-valued function on $\mathbb{R}^d$.

Equipped with the $K \times 1$ vector $\mathbf{V}$ and the $K \times J$ matrix $\Xi$, we solve the regression problem
\begin{equation}
\label{eqn:sindy}
\mathbf{V} = \Xi \bbeta + \epsilon
\end{equation}
for the $J \times 1$ vector $\bbeta$ using an iteratively thresholded least-squares algorithm.  The algorithm has one constant hyperparameter, $\lambda > 0$.  The algorithm is then succinctly described as follows: (i) estimate $\bbeta$ using ordinary least squares, and then (ii) reset to zero all components of $\bbeta$ that are less than the threshold $\lambda$.  Once components of $\bbeta$ are reset to zero, they stay frozen at zero.  We then repeat steps (i) and (ii) until $\bbeta$ stabilizes to its converged value.

As shown recently  \cite{ZhangSchaeffer2018}, this algorithm converges in a finite number of steps to an approximate minimizer of $\| \mathbf{V} - \Xi \bbeta \|^2 + \lambda^2 \| \bbeta \|_0 $.  Here $\| \bbeta \|_0$ denotes the number of nonzero entries of $\bbeta$.  Hence, increasing the parameter $\lambda$ leads to a more sparse set of coefficients $\bbeta$.  Once we have fit the regression model in this way, we obtain an interpretable model of $\widehat{V}$, specifically:
\begin{equation}
\label{eqn:intmod}
\widehat{V}(\mathbf{x}) = \sum_{j=1}^{J} \beta_j \xi^j(\mathbf{x}) + \epsilon.
\end{equation}
If $\bbeta$ is highly sparse, most of the coefficients $\beta_j$ will be zero.  Suppose that the candidate functions $\xi^j$ are well-known functions such as positive or negative powers of the coordinates $x_i$ of the input $\mathbf{x}$.  In this case, the right-hand side will be a relatively short algebraic expression that is as interpretable as most potential energy functions routinely encountered in classical physics.  The norm of $\epsilon$ here captures the error in this sparse approximation of $\widehat{V}$.  In general, one chooses $\lambda$ to balance the sparsity of $\bbeta$ with the quality of the approximation $\| \epsilon \|$.
\section{Tests}
We now describe a series of increasingly complex tests that demonstrate the proposed method.  For each such model, we use either exact solutions or fine-scale numerical integration to create a corpus of time series measurements.  Using the time series, we train a neural potential energy model, which we then interpret using SINDy.  We use NumPy/SciPy or Mathematica for all data generation, TensorFlow for all neural network model development/training, and the \texttt{sindyr} package \cite{DaleBhatSindyr} in R to interpret the neural potential.  In what follows, the mass matrix $M$ in (\ref{eqn:ham}) is the identity unless specified otherwise.  In all cases, we train the neural network using gradient descent.  We are committed to releasing all code/data at \url{https://github.com/hbhat4000/learningpotentials}.

\subsection{Simple Harmonic Oscillator}
The first model we consider is the simple harmonic oscillator ($d=1$) with Hamiltonian
\begin{equation}
\label{eqn:shoHam}
H(q,p) = \frac{p^2}{2} + \frac{q^2}{2}.
\end{equation}
Exact trajectories consists of circles centered at the origin in $(q,p)$ space.  For training data, we use $R=10$ such circles; for $1 \leq i \leq 10$, the $i$-th circle passes through an initial condition $(q(0),p(0)) = (0, i)$.  We include $N=1000$ steps of each trajectory, recorded with a time step of $0.01$, in the training set.  Here our goal is to check how closely the neural potential $\widehat{V}(q)$ can track the true potential $V(q) = q^2/2$.  We take the neural potential model to have two hidden layers, each with $16$ units and $\tanh$ activations.  We train for $50000$ steps at a learning rate of $0.01$.

\begin{figure}[t]
\begin{center}
\includegraphics[width=3in]{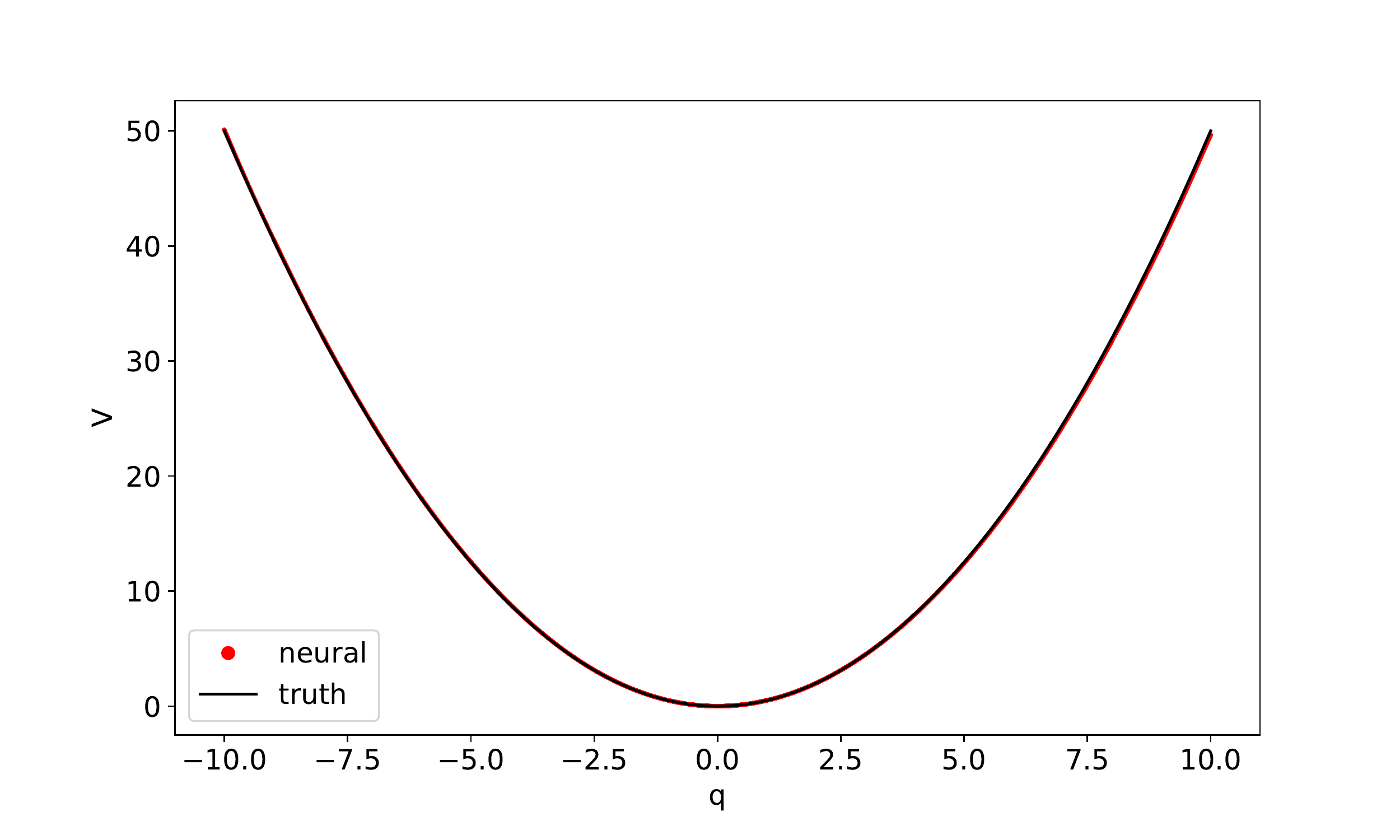}
\end{center}
\caption{For the simple harmonic oscillator (\ref{eqn:shoHam}), after adjusting a constant bias, the neural potential $\widehat{V}(q)$ closely matches the true potential $V(q) = q^2/2$.}
\label{fig:shoplot}
\end{figure}

In Figure \ref{fig:shoplot}, we plot both the trained neural potential $\widehat{V}$ (in red) and the true potential $V(q) = q^2/2$ (in black).  When plotting $\widehat{V}$, we have subtracted a constant bias (the minimum obtained value of $\widehat{V}$) so that the curve reaches a minimum value of zero.  Note that this constant bias is completely unimportant for physics, as only $\nabla V$ appears in Hamilton's equations (\ref{eqn:ham}) and the loss (\ref{eqn:risk}).  However, the constant bias does lead us to include an intercept in the regression model (\ref{eqn:sindy}), i.e., to include $1$ in the set of candidate functions for SINDy.  We follow this practice in all uses of SINDy below.

We apply SINDy to the learned potential $\widehat{V}(q)$ with candidate functions $\{1, q, q^2, q^3\}$.  In the following test, and in fact throughout this paper, we start with $\lambda = 1$ and tune $\lambda$ downward until the error $\| \epsilon \|$---between the neural network potential  $\widehat{V}(q)$ and the SINDy-computed approximation---drops below $10^{-10}$.  We find that with $\lambda = 0.04$, the estimated system is
\begin{equation}
\label{eqn:sindysho}
\widehat{V}(q) \approx \beta_0 + \beta_2 q^2
\end{equation}
with $\beta_0 \approx -49.18$ and $\beta_2 \approx 0.4978$. We see that $\widehat{V}$ closely tracks the true potential $V(q) = q^2/2$ up to the constant bias term, which can be ignored.

\begin{figure}[t]
\begin{center}
\includegraphics[width=3in]{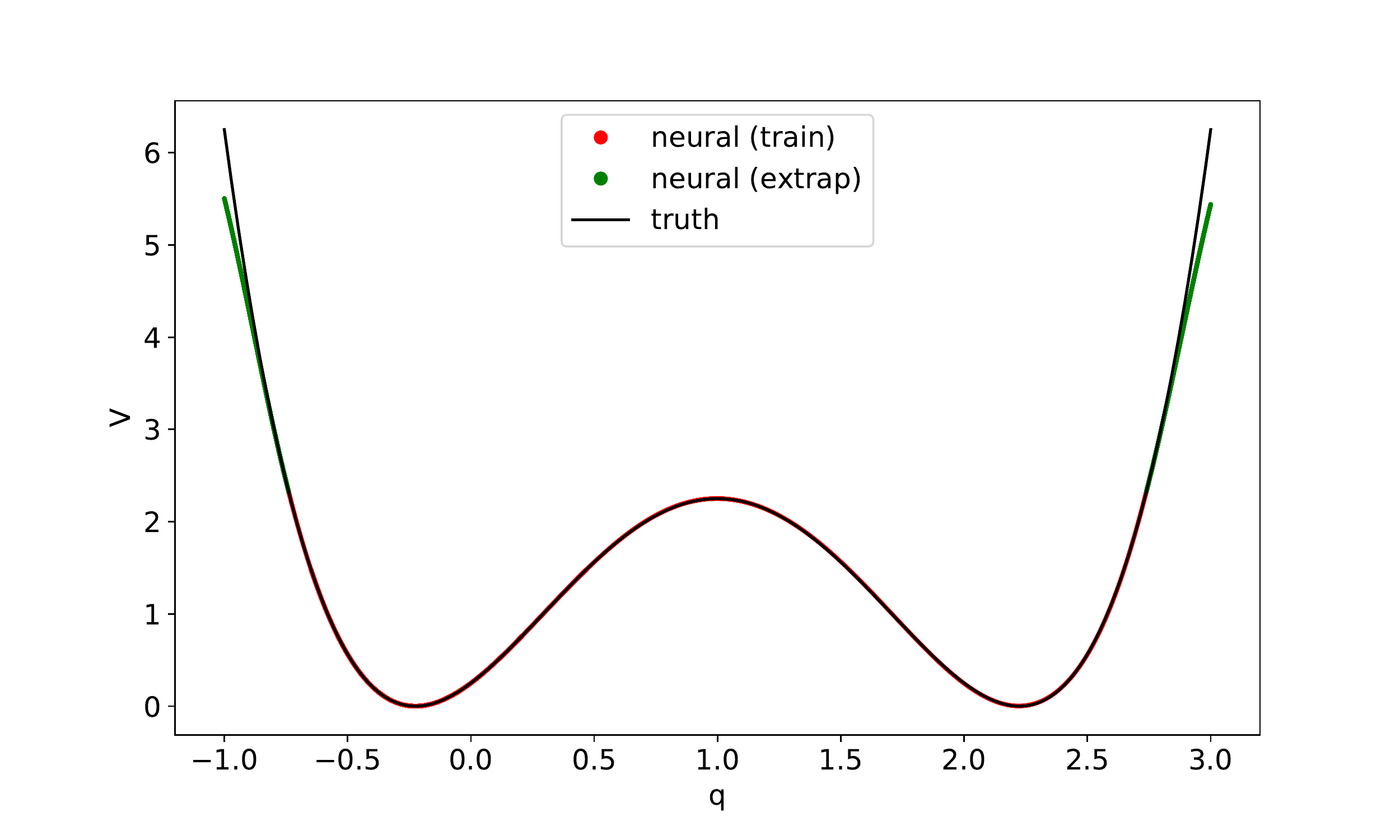}
\end{center}
\caption{For the double well potential (\ref{eqn:dblwell}), the neural potential $\widehat{V}(q)$ trained on $\mathcal{T}_1$ closely matches the true potential $V(q)$.  This training set includes one high-energy trajectory that visits both wells.  In red, we plot $V(q)$ for $q \in \mathcal{T}_1$; in green, we plot $V(q)$ for $q \in [-1,3] \setminus \mathcal{T}_1$. Potentials were adjusted by a constant bias so that they both have minimum values equal to zero.}
\label{fig:dblwell}
\end{figure}

\begin{figure}[t]
\begin{center}
\includegraphics[width=2.25in]{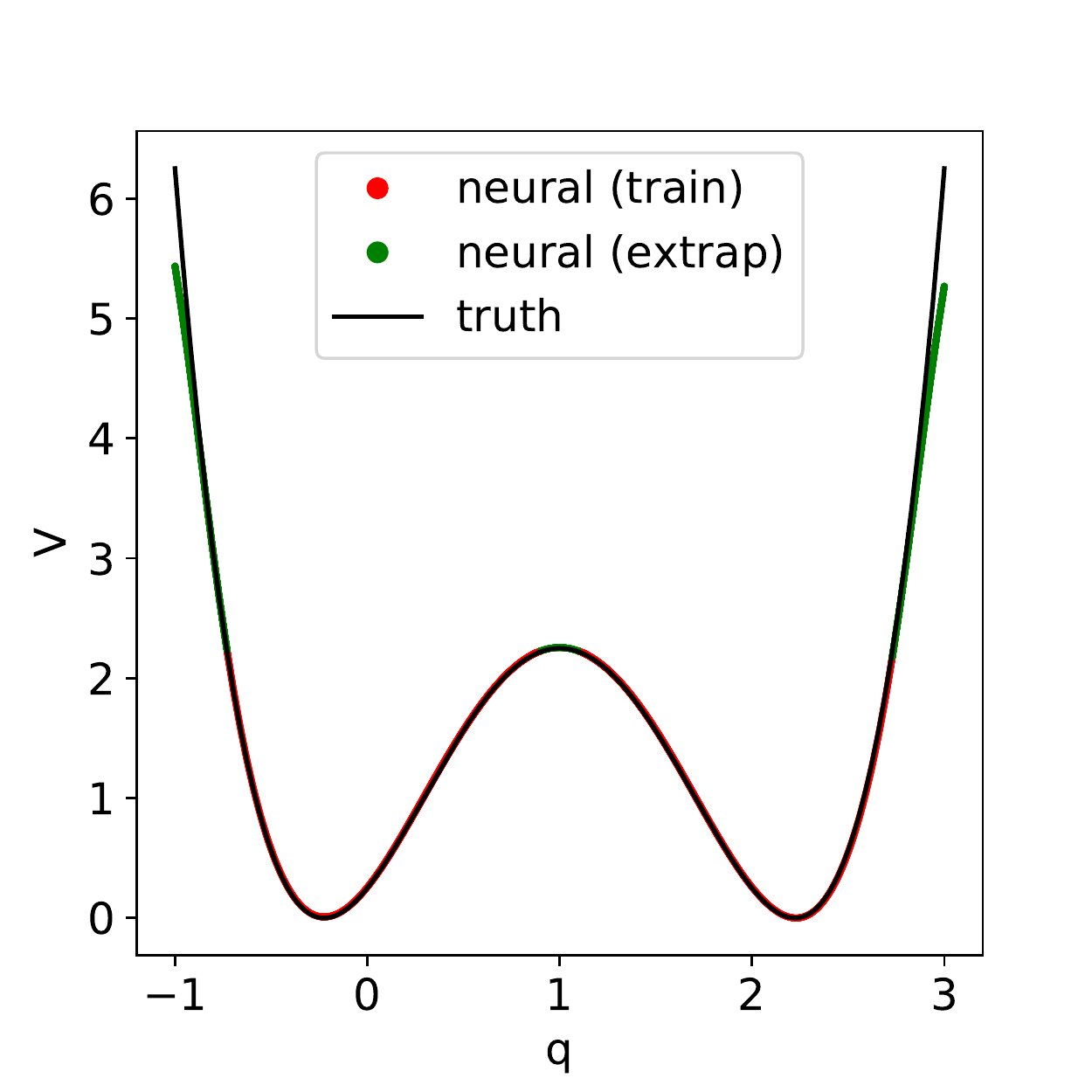} \hfill
\includegraphics[width=2.25in]{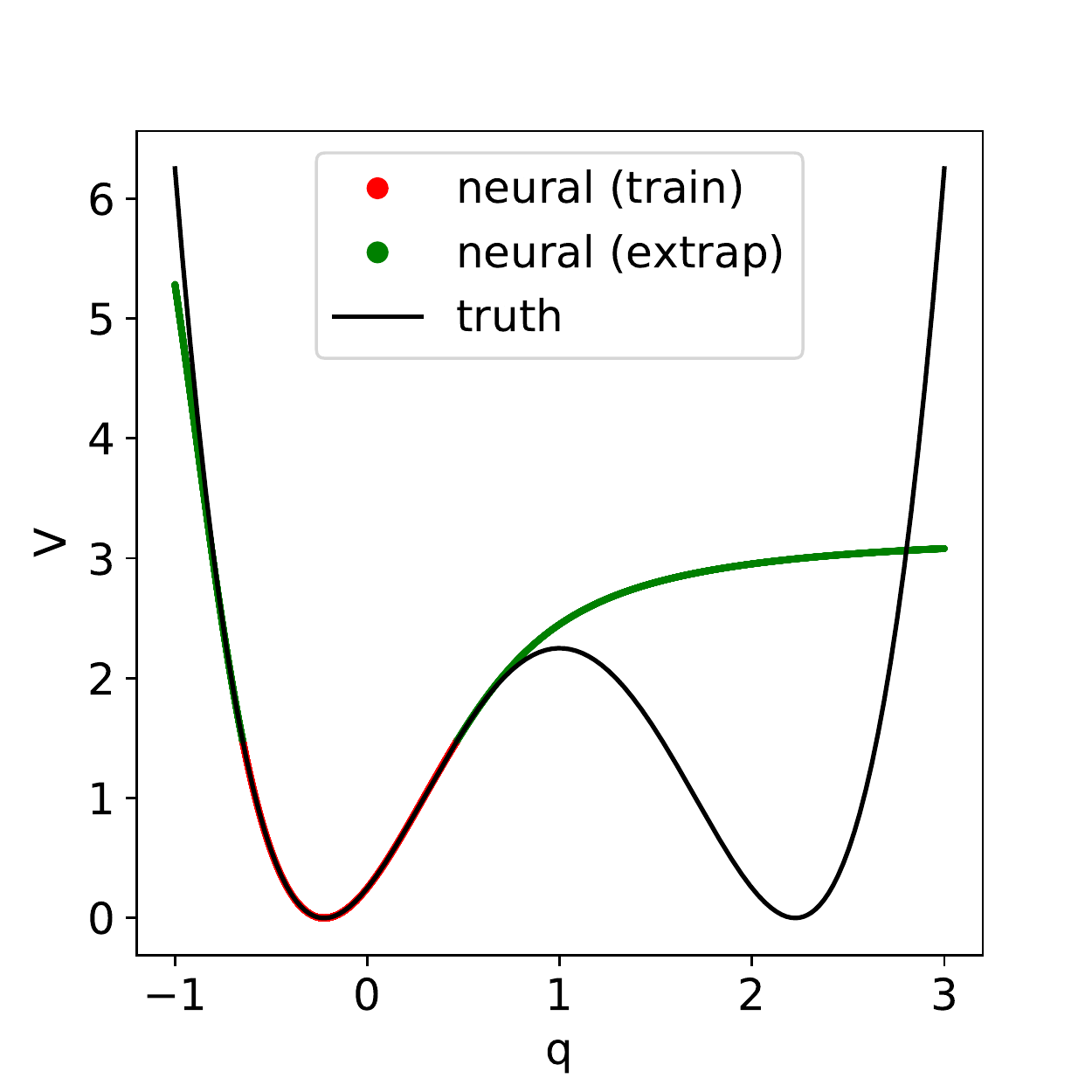}
\end{center}
\caption{For the double well potential (\ref{eqn:dblwell}), we train neural potentials $\widehat{V}(q)$ using, in turn, the training sets $\mathcal{T}_2$ (left) and $\mathcal{T}_3$ (right).  We plot in red $\widehat{V}(q)$ only for the values of $q$ covered by the respective training sets; in green, we extrapolate $\widehat{V}(q)$ to values of $q$ that are not in the respective training sets.  Since $\mathcal{T}_2$ includes two trajectories, one from each well, the neural potential captures and extrapolates well to both wells.  Conversely, because $\mathcal{T}_3$ only includes trajectories that stay in one well, the neural potential completely misses one well.  Potentials were adjusted by a constant bias so that they all have minimum values equal to zero.}
\label{fig:dblwell2}
\end{figure}

\subsection{Double Well}
Let us consider a particle in a double well potential ($d=1$)
\begin{equation}
\label{eqn:dblwell}
V(q) = x^2 (x-2)^2 - (x-1)^2.
\end{equation}
We take the kinetic energy to be $T(p) = p^2/2$.    We now use explicit Runge-Kutta integration in Mathematica to form three training sets:
\begin{itemize}
\item Training set $\mathcal{T}_1$ includes $R=10$ trajectories with random initial conditions $(q(0), p(0))$ chosen uniformly on $[-1,1]^2$, one of which has sufficiently high energy to visit both wells.
\item Training set $\mathcal{T}_2$ consists of $R=2$ trajectories, each of which starts and stays in an opposing well.  The first trajectory has initial condition $(q(0),p(0)) = (3,0)$ while the second has initial condition $(q(0),p(0)) = (-1,0)$.  These $q(0)$ values are symmetric across $q=1$, the symmetry axis of $V(q)$.
\item Training set $\mathcal{T}_3$ has only $R=2$ trajectories that stay in the left well only.
\end{itemize}
For each trajectory, we record $5001$ points at a time step of $0.001$.  We take the neural potential model to have two hidden layers, each with $16$ units and $\tanh$ activations.  For each training set $\mathcal{T}_m$, we train for $50000$ steps at a learning rate of $0.01$.

We seek to understand how the choice of training set $\mathcal{T}_m$ affects the ability of the neural potential $\widehat{V}$ to track the true potential (\ref{eqn:dblwell}). We plot and discuss the results in Figures \ref{fig:dblwell} and \ref{fig:dblwell2}.  Overall, the neural potentials trained using $\mathcal{T}_1$ and $\mathcal{T}_2$ match $V(q)$ closely---both on the training set and extrapolated to the rest of the interval $-1 \leq q \leq 3$.  Clearly, the neural potential trained using $\mathcal{T}_3$ only captures one well and does not extrapolate correctly to the rest of the domain.

Let $\widehat{V}^m(q)$ denote the neural potential trained on $\mathcal{T}_m$. We now apply SINDy to the output of each $\widehat{V}^m$ only on its respective training set $\mathcal{T}_m$, with candidate functions $\{1, q, q^2, q^3, q^4, q^5, q^6 \}$.  For reference, the ground truth $V(q)$ can be written as
$V(q) = -1 + 2 q + 3 q^2 - 4 q^3 + q^4$.
Adjusting $\lambda$ downward as described above, we find with $\lambda = 0.5$ the following algebraic expressions:
\begin{align*}
\widehat{V}^{1}(q) &\approx -8.138 + 2.0008 q + 3.0009 q^2 - 4.0009 q^3 + 1.0001 q^4 \\
\widehat{V}^{2}(q) &\approx -6.061 + 2.0032 q + 3.0054 q^2 - 4.0148 q^3 + 0.9748 q^4 \\
\widehat{V}^{3}(q) &\approx -6.165 + 1.9909 q + 2.9991 q^2 - 3.9886 q^3 + 0.9955 q^4
\end{align*}
Noting that the constant terms are irrelevant, we note here that \emph{all three models agree closely with the ground truth}.  The agreement between $V$ and the algebraic forms of $\widehat{V}^1$ and $\widehat{V}^2$ was expected.  We find it somewhat surprising that SINDy, when applied to the output of $\widehat{V}^3$ on its training set $\mathcal{T}^3$, yields a quartic polynomial with two wells.

\subsection{Central Force Problem}
We consider a central force problem for one particle ($d=3$) with Hamiltonian
\begin{equation}
\label{eqn:cfHam}
H(\q,\p) = \frac{ \| \p \|^2}{ 2} + \| \q \|^{-1} + (10 - \| \q \|)^{-1}.
\end{equation}
The norm here is the standard Euclidean norm.  Using explicit Runge-Kutta integration in Mathematica, we generate $R=1$ trajectory with random initial condition $(\q(0), \p(0))$ chosen uniformly on $[-1,1]^6$.  Using this trajectory, we compute $r(t) = \| \q(t) \|$ as well as $\dot{r}(t) = dr/dt$.  We save the $(r(t),\dot{r}(t))$ trajectories at $N = 20001$ points with a time step of $0.001$.  We then search for a reduced-order ($d=1$) model with Hamiltonian
\begin{equation}
\label{eqn:romHam}
H(r,\dot{r}) = \frac{ \dot{r}^2 }{2} + \widetilde{V}(r),
\end{equation}
where $\widetilde{V}(r)$ is a neural potential.   We take the neural potential model to have two hidden layers, each with $16$ units.  We train for $500000$ steps at a learning rate of $10^{-3}$, first using exponential linear unit activations $\psi(x) = \begin{cases} x & x \geq 0 \\ \exp(x) - 1 & x < 0. \end{cases}$. We then initialize the neural network using the learned weights/biases and retrain using softplus activations $\phi(x) = \log(1 + \exp(x))$---this activation was chosen to enable series expansions of $\widehat{V}(r)$, as described in greater detail below.  Prior to retraining, we also change the network by adding an exponential function to the output layer---we incorporate this function to better model the steep gradients in the potential near $r=0$ and $r=10$.  When we retrain, we take $500000$ steps at a learning rate of $10^{-3}$.  We carry out the training in two stages because training directly with softplus activations and exponential output failed.

\begin{figure}[t]
\begin{center}
\includegraphics[width=3in]{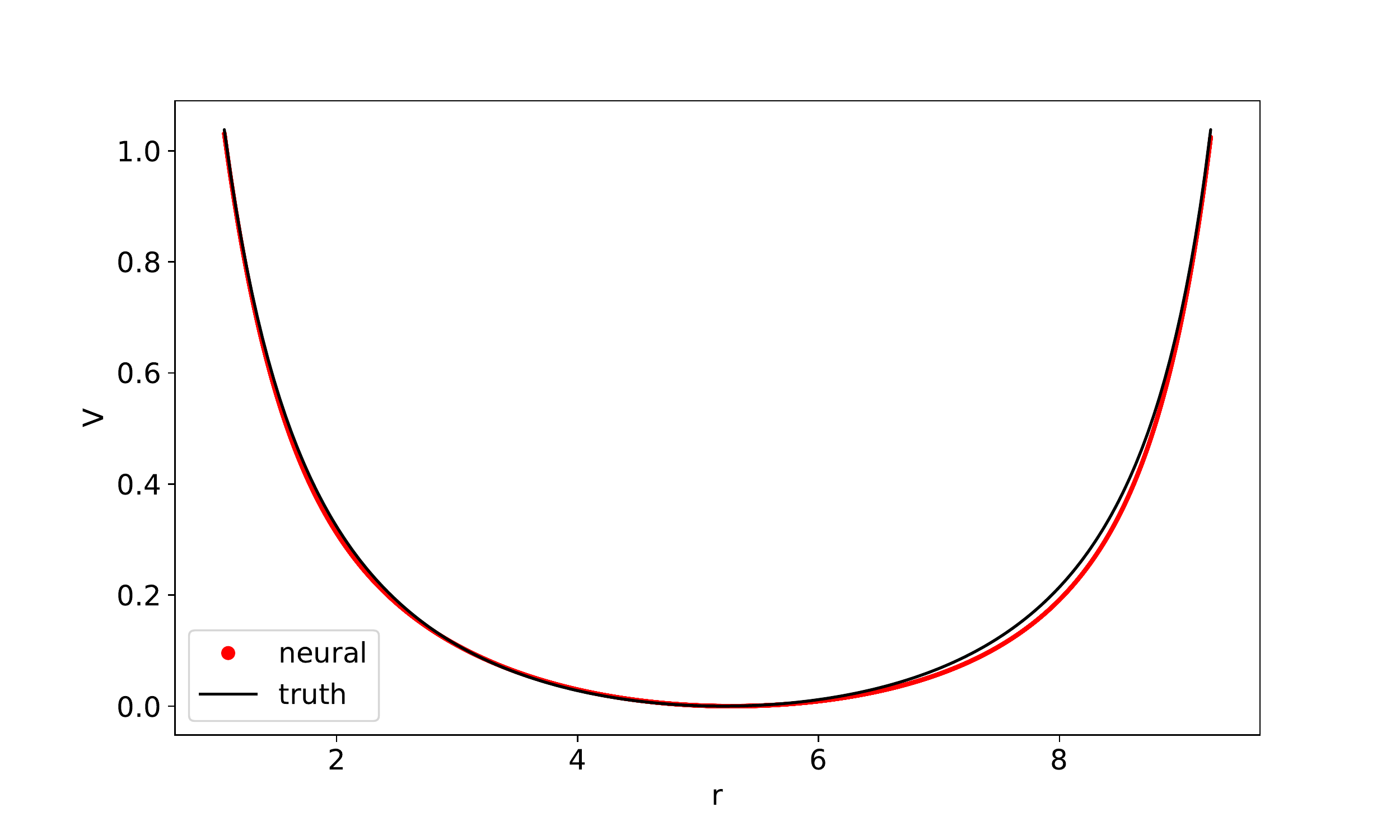}
\end{center}
\caption{For the central force problem (\ref{eqn:romHam}), after adjusting a constant bias, the neural potential $\widehat{V}(r)$ closely matches the effective potential $V_\text{eff}(r)$.}
\label{fig:romplot}
\end{figure}

For this problem, classical physics gives us an \emph{effective potential}
\begin{equation}
\label{eqn:effpot}
V_\text{eff}(r) = r^{-1} + (10-r)^{-1} + \ell^2/(2 r^2).
\end{equation}
where $\ell$ is a conserved quantity determined from the initial condition.  In Figure \ref{fig:romplot}, we plot both the trained neural potential $\widehat{V}$ (in red) and the effective potential $V_\text{eff}(r)$ (in black).  After adjusting for the constant bias term, we find close agreement.

We then exported the weight and bias matrices to Mathematica, forming the neural potential model
\begin{equation}
\label{eqn:neuralpot}
\widehat{V}(r) = W_3 \phi( W_2 \phi(W_1 r + b_1) + b_2) + b_3.
\end{equation}
Unlike $\psi$, the softplus activation $\phi$ is amenable to series expansion via symbolic computation.   In particular, since we can see that the effective potential $V_\text{eff}(r)$ is a rational function, we explored Pad\'{e} expansions of $\widehat{V}(r)$.  These attempts were unsuccessful in the sense that we did not obtain models of $\widehat{V}(r)$ that are any more interpretable than the compositional form of (\ref{eqn:neuralpot}). 

Turning to SINDy, we formed a library of candidate functions
$$
\{1, r^{-1}, r^{-2}, r^{-3}, (10-r)^{-1}, (10-r)^{-2}, (10-r)^{-3}\}.
$$
Adjusting $\lambda$ in the same manner described above, we find that with $\lambda = 0.15$, the estimated model is
\begin{equation}
\label{eqn:sindyrom}
\widehat{V}(r) \approx \beta_0 + \beta_1 r^{-1} + \beta_2 r^{-2} + \beta_4 (10-r)^{-1}
\end{equation}
Here $\beta_0 \approx -0.2384$, $\beta_1 \approx 1.005$, $\beta_2 \approx 0.4461$, and $\beta_4 \approx 0.9723$.  We see from the form of $V_\text{eff}(r)$ given above that $\beta_1$ and $\beta_4$ are both close to the ground truth values of $1$.  Note that for the trajectory on which the system was trained, we have $\ell^2 /2 \approx 0.4655$.  Hence $\beta_2$ has an error of less than $4.2\%$.  This demonstrates a successful application of SINDy to interpret the neural potential as a rational function; this interpretation of $\widehat{V}$ is itself close to $V_\text{eff}$.

\subsection{Charged Particles in Coulomb Potential}
We now consider two oppositely charged particles ($d=6$) subject to the classical Coulomb electrostatic potential.   We take the mass matrix to be $M = \operatorname{diag}(1, 1/2)$.  The kinetic energy is $T(\p) = \p^T M^{-1} \p /2$.  If we partition $\q = (\q_1, \q_2)$ where $\q_i$ is the position of the $i$-th particle, then the potential is
\begin{equation}
\label{eqn:coulPot}
V(\q) = -\frac{1}{4 \pi} \frac{1}{\| \q_1 - \q_2\|}.
\end{equation}
Here we apply the St\"ormer-Verlet algorithm, a symplectic method, to generate $R = 1000$ trajectories, each with $N = 10001$ points recorded at a time step of $0.001$.  Each trajectory starts with random initial conditions $(\q(0), \p(0))$ chosen from a standard normal.  For this problem, our goal is to use the data to recover $V$.  We train two different neural potential models with increasing levels of prior domain knowledge:
\begin{enumerate}
\item We first set up the neural network's input layer to compute from $\q$ the difference $\q_1 - \q_2 \in \mathbb{R}^3$; the neural potential then transforms this three-dimensional input into a scalar output.  The neural network here has $8$ hidden layers, each with $16$ units and $\tanh$ activations.  Using only $800$ of the $R = 1000$ trajectories, we first train using the first $100$ points from each of the $800$ trajectories, taking $500000$ steps at a learning rate of $0.01$. Again restricting ourselves to the $800$ training trajectories, we then use the next $100$ points, followed by the next $100$ points, etc., each time taking $500000$ steps at a learning rate of $0.01$.  As the training loss was observed to be sufficiently small ($\approx 0.006021$), we halted training.

\begin{figure}[t]
\begin{center}
\includegraphics[width=2.25in]{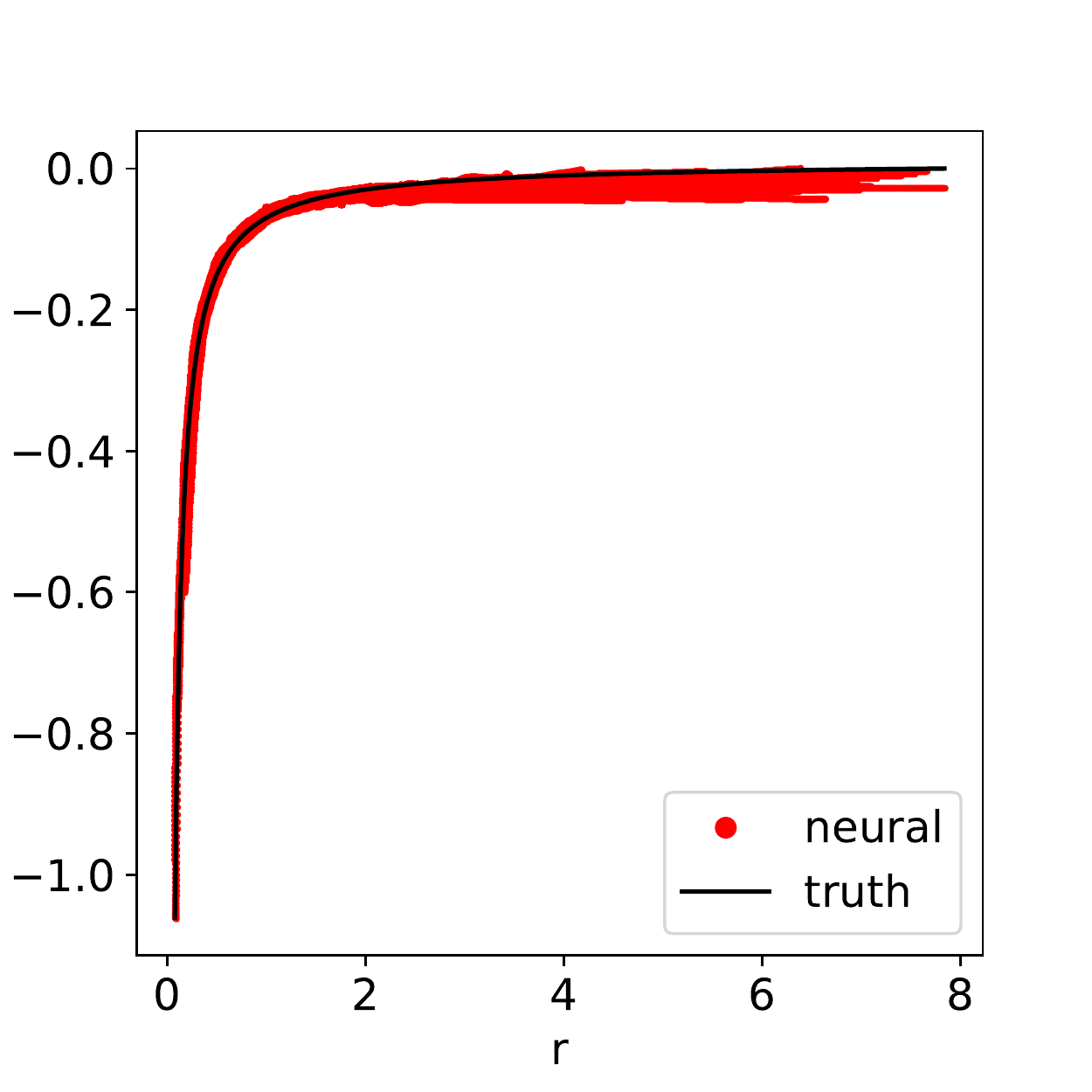} \hfill
\includegraphics[width=2.25in]{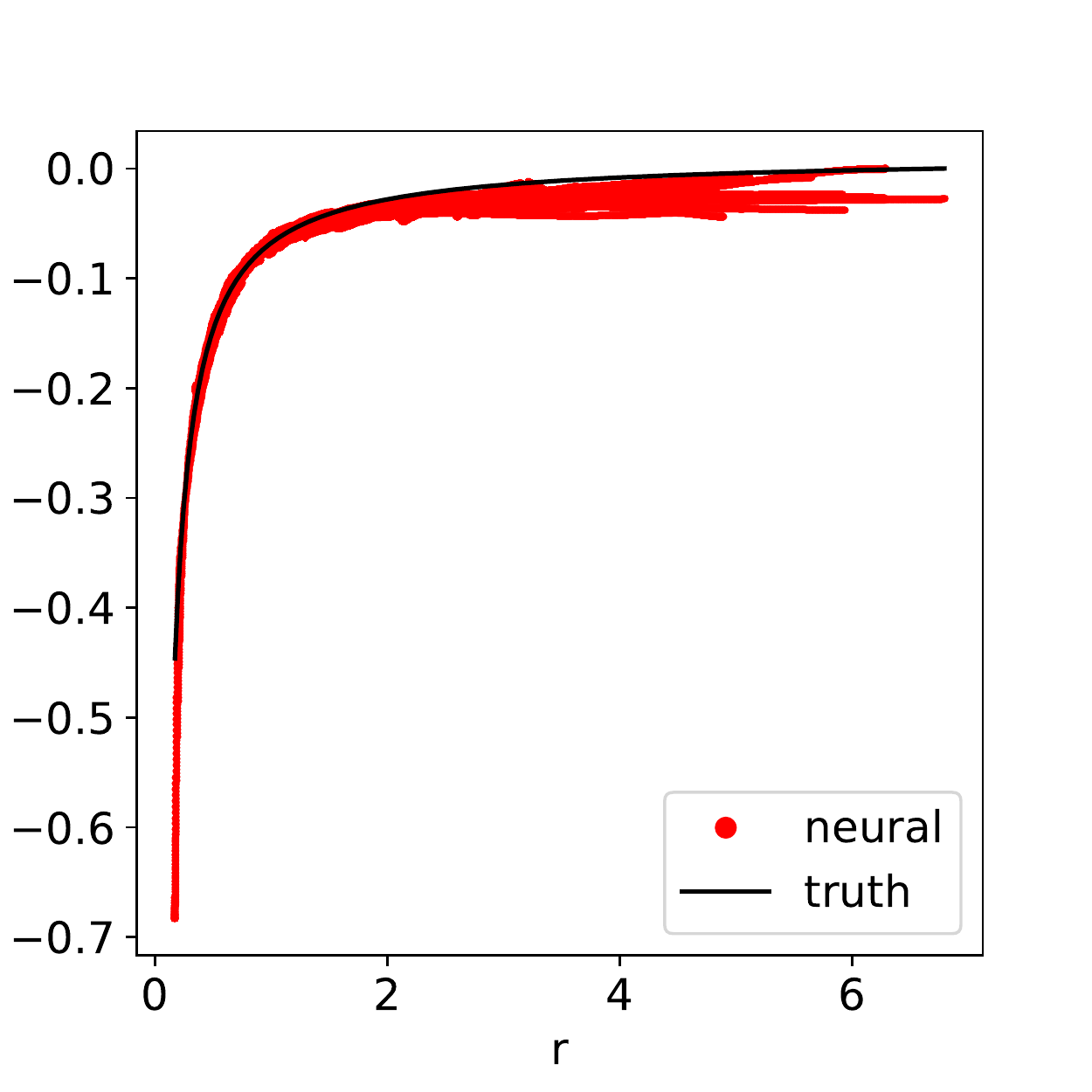}
\end{center}
\caption{Here we plot both training (left) and test (right) results for the Coulomb problem (\ref{eqn:coulPot}).    For both plots, we have subtracted a constant bias, the maximum value of the neural potential on the data set in question.  These results are for a neural potential $\widehat{V}$ that is a function of the difference $\q_1 - \q_2$ between the two charged particles' positions; for each $\q$ in the training and test sets, we plot $\widehat{V}(\q_1 - \q_2)$ versus $r = \| \q_1 - \q_2\|$.  We also plot the true potential (\ref{eqn:coulPot}) versus $r$.  Both training and test plots show reasonable agreement between the neural potential and the ground truth.}
\label{fig:pot3}
\end{figure}

In Figure \ref{fig:pot3}, we plot both training (left) and test (right) results.  The training results are plotted with the first $1000$ points of the $800$ trajectories used for training, while the test results are plotted with the first $1000$ points of the $200$ held out trajectories.  For both plots, we subtracted the maximum computed value of $\widehat{V}$ (on each respective data set).  In each plot, we plot $\widehat{V}$ (on all points $\q$ in the training and test sets) versus $r = \| \q_1 - \q_2\|$.

Overall, we see reasonable agreement between the neural potential and the ground truth.  Note that the neural network is essentially tasked with discovering that it should compute the inverse of the norm of  $\q_1 - \q_2$.  We suspect that this function of $\q_1 - \q_2$ may be somewhat difficult to represent using a composition of activation functions and linear transformations as in (\ref{eqn:neuralpot}).  Despite training for a large number of steps, there is noticeable variation in neural potential values for large $r$.

We now apply SINDy to $\widehat{V}$ (on the training set) using candidate functions $\{1, r^{-1}, r^{-2}, r^{-3}\}$.  Adjusting $\lambda$ as described above, we find with $\lambda = 0.04$, the approximation
\begin{equation}
\label{eqn:sindypot3}
\widehat{V}(r) \approx \beta_0 + \beta_1 r^{-1}
\end{equation}
with $\beta_0 \approx 0.7602$ and $\beta_1 \approx -0.06911$.  For comparison, the ground truth coefficient of $r^{-1}$ is $-(4 \pi)^{-1} \approx -0.07958$.

\item We then rearchitect the network to include a layer that takes the input $\q$ and computes the norm of the difference $\| \q_1 - \q_2\|$; the rest of the neural potential is then a scalar function of this scalar input.  Here the neural network has $8$ hidden layers, each with $8$ units and $\tanh$ activations.  We train for $50000$ steps with learning rate of $0.05$.  Note that here, for training, we use $N = 5001$ time steps of only $100$ trajectories.

\begin{figure}[t]
\begin{center}
\includegraphics[width=2.25in]{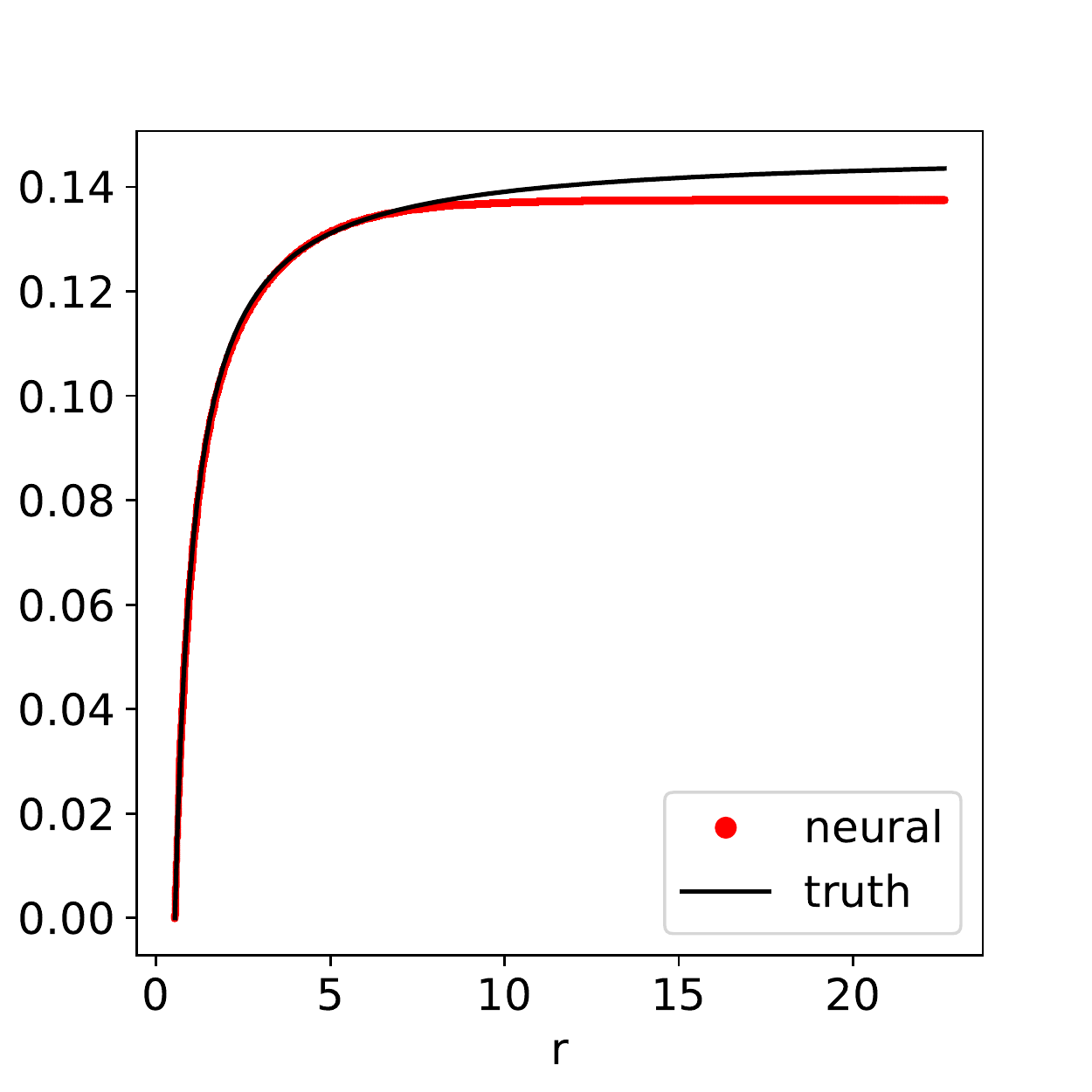} \hfill
\includegraphics[width=2.25in]{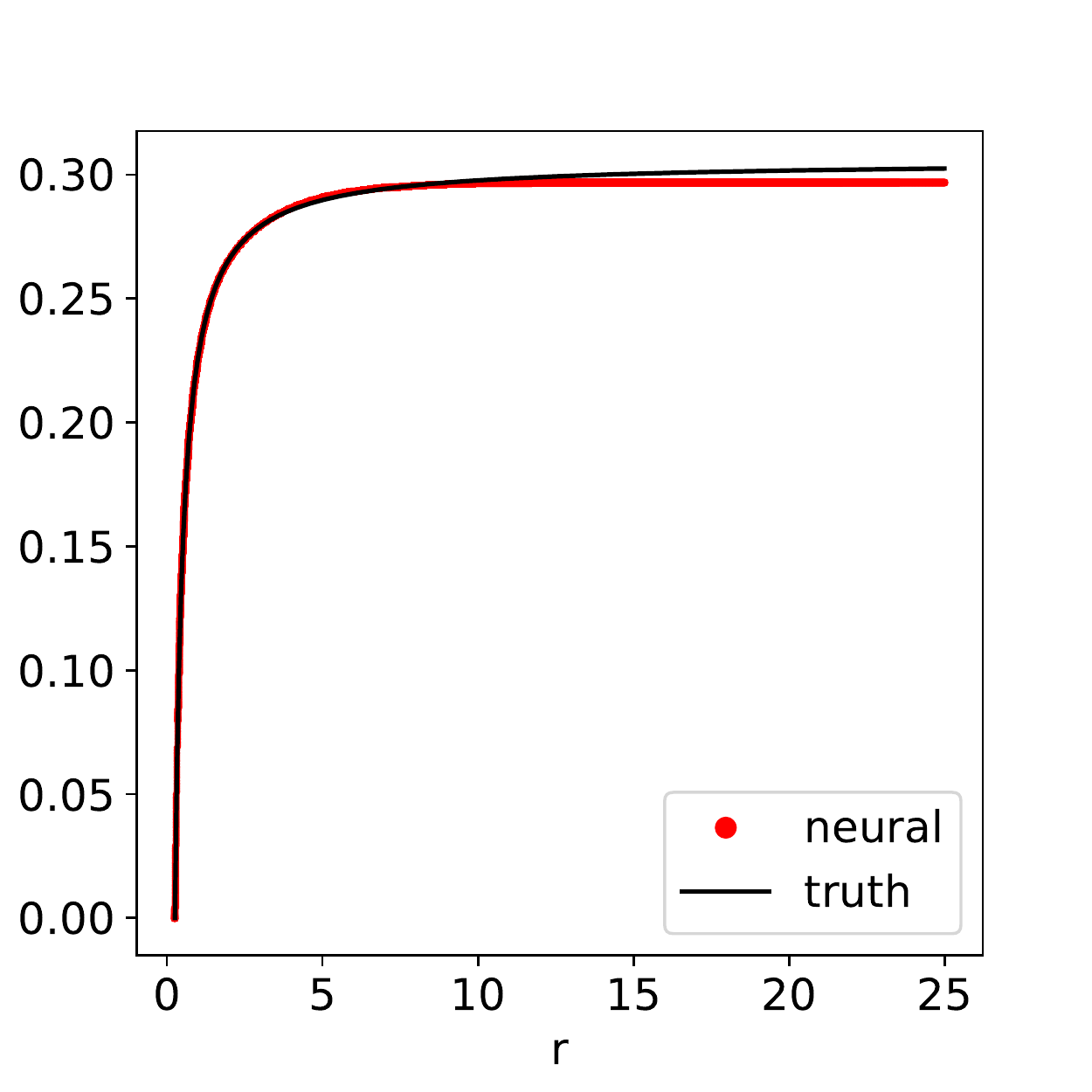}
\end{center}
\caption{Here we plot both training (left) and test (right) results for the Coulomb problem (\ref{eqn:coulPot}).    For both plots, we have subtracted a constant bias, the maximum value of the neural potential on the data set in question.  These results are for a neural potential $\widehat{V}$ that is a function of the distance $r = \| \q_1 - \q_2 \|$ between the two charged particles; for each $\q$ in the training and test sets, we plot $\widehat{V}(r)$ versus $r = \| \q_1 - \q_2\|$.  We also plot the true potential (\ref{eqn:coulPot}) versus $r$.  Both training and test plots show excellent agreement between the neural potential and the ground truth.}
\label{fig:pot2}
\end{figure}

In Figure \ref{fig:pot2}, we plot both training (left) and test (right) results.  The training results are plotted with the first $5001$ points of the $100$ trajectories used for training, while the test results are plotted with a completely different set of $100$ trajectories, each of length $5001$  For both plots, we subtracted the minimum computed value of $\widehat{V}$ (on each respective data set).  In each plot, we compute $\widehat{V}$ on all points $\q$ in the training and test sets, and then plot these $\widehat{V}$ values versus $r = \| \q_1 - \q_2\|$.

To generate an interpretable version of $\widehat{V}$ (on the training set), we apply SINDy with candidate functions $\{1, r^{-1}, r^{-2}, r^{-3}\}$.  Adjusting $\lambda$ as described above, we find with $\lambda = 0.05$, the approximation
\begin{equation}
\label{eqn:sindypot2}
\widehat{V}(r) \approx \beta_0 + \beta_1 r^{-1}
\end{equation}
with $\beta_0 \approx 2.267$ and $\beta_1 \approx -0.07792$.  This computed value of $\beta_1$ is less than $2.1\%$ away from the ground truth value of $-(4 \pi)^{-1}$; the error for the earlier approximation (\ref{eqn:sindypot3}) was just over $13.1\%$.

Incorporating prior knowledge that the potential should depend only on $r$ dramatically improves the quality of the learned potential.  Essentially, we have eliminated the need for the neural network to learn the norm function.  We outperform the results from Figure \ref{fig:pot3} using a less complex network, trained for fewer steps and a larger learning rate.  Comparing with Figure \ref{fig:pot3}, we see that Figure \ref{fig:pot2} features reduced variation in $\widehat{V}(r)$ for large $r$, and improved test set results as well.
\end{enumerate}

\section{Conclusion}
We conclude that, for the examples we have explored, our approach does lead to accurate potentials that can themselves be approximated closely by interpretable, closed-form algebraic expressions.
In ongoing/future work, we plan to apply the techniques described here to high-dimensional systems for which reduced-order (i.e., effective) potentials are unknown.  We also seek to extend our method to quantum Hamiltonian systems.  While we have focused here on clean data from known models, we are also interested in learning potentials from noisy time series.  We expect that by adapting the method of \cite{BhatICML2019}, we will be able to simultaneously filter the data and estimate an interpretable neural potential.

%
%
%
\bibliographystyle{splncs04}

\begin{thebibliography}{10}
\providecommand{\url}[1]{\texttt{#1}}
\providecommand{\urlprefix}{URL }
\providecommand{\doi}[1]{https://doi.org/#1}

\bibitem{Artrith2016}
Artrith, N., Urban, A.: An implementation of artificial neural-network
  potentials for atomistic materials simulations: Performance for {TiO}2.
  Computational Materials Science  \textbf{114},  135--150 (Mar 2016).
  \doi{10.1016/j.commatsci.2015.11.047}

\bibitem{Behler2016}
Behler, J.: Perspective: Machine learning potentials for atomistic simulations.
  The Journal of Chemical Physics  \textbf{145}(17),  170901 (2016).
  \doi{10.1063/1.4966192}

\bibitem{Behler07}
Behler, J., Parrinello, M.: Generalized neural-network representation of
  high-dimensional potential-energy surfaces. Phys. Rev. Lett.  \textbf{98},
  146401 (Apr 2007). \doi{10.1103/PhysRevLett.98.146401}

\bibitem{bhat_nonparametric_2016}
Bhat, H.S., Madushani, R.W.M.A.: Nonparametric {Adjoint}-{Based} {Inference}
  for {Stochastic} {Differential} {Equations}. In: 2016 {IEEE} {International}
  {Conference} on {Data} {Science} and {Advanced} {Analytics} ({DSAA}). pp.
  798--807 (2016). \doi{10.1109/DSAA.2016.69}

\bibitem{Brunton2016}
Brunton, S.L., Proctor, J.L., Kutz, J.N.: Discovering governing equations from
  data by sparse identification of nonlinear dynamical systems. Proceedings of
  the National Academy of Sciences  \textbf{113}(15),  3932--3937 (2016)

\bibitem{BhatDale2018}
Dale, R., Bhat, H.S.: Equations of mind: Data science for inferring nonlinear
  dynamics of socio-cognitive systems. Cognitive Systems Research  \textbf{52},
   275--290 (2018)

\bibitem{DaleBhatSindyr}
Dale, R., Bhat, H.S.: sindyr: Sparse Identification of Nonlinear Dynamics
  (2018), \url{https://CRAN.R-project.org/package=sindyr}, r package version
  0.2.1

\bibitem{Duncker2019}
Duncker, L., Bohner, G., Boussard, J., Sahani, M.: Learning interpretable
  continuous-time models of latent stochastic dynamical systems. In: Chaudhuri,
  K., Salakhutdinov, R. (eds.) Proceedings of the 36th International Conference
  on Machine Learning. Proceedings of Machine Learning Research, vol.~97, pp.
  1726--1734. PMLR, Long Beach, California, USA (09--15 Jun 2019),
  \url{http://proceedings.mlr.press/v97/duncker19a.html}

\bibitem{Hansen15}
Hansen, K., Biegler, F., Ramakrishnan, R., Pronobis, W., von Lilienfeld, O.A.,
  Muller, K.R., Tkatchenko, A.: Machine learning predictions of molecular
  properties: Accurate many-body potentials and nonlocality in chemical space.
  The Journal of Physical Chemistry Letters  \textbf{6}(12),  2326--2331
  (2015). \doi{10.1021/acs.jpclett.5b00831}

\bibitem{Ramakrishnan15}
Ramakrishnan, R., Hartmann, M., Tapavicza, E., von Lilienfeld, O.A.: Electronic
  spectra from {TDDFT} and machine learning in chemical space. The Journal of
  Chemical Physics  \textbf{143}(8),  084111 (2015). \doi{10.1063/1.4928757}

\bibitem{BhatICML2019}
Raziperchikolaei, R., Bhat, H.S.: A block coordinate descent proximal method for
  simultaneous filtering and parameter estimation. In: Chaudhuri, K.,
  Salakhutdinov, R. (eds.) Proceedings of the 36th International Conference on
  Machine Learning. Proceedings of Machine Learning Research, vol.~97, pp.
  5380--5388. PMLR, Long Beach, California, USA (09--15 Jun 2019),
  \url{http://proceedings.mlr.press/v97/raziperchikolaei19a.html}

\bibitem{Sahoo2018}
Sahoo, S., Lampert, C., Martius, G.: Learning equations for extrapolation and
  control. In: Dy, J., Krause, A. (eds.) Proceedings of the 35th International
  Conference on Machine Learning. Proceedings of Machine Learning Research,
  vol.~80, pp. 4442--4450. PMLR, Stockholmsmässan, Stockholm Sweden (10--15
  Jul 2018), \url{http://proceedings.mlr.press/v80/sahoo18a.html}

\bibitem{ZhangSchaeffer2018}
{Zhang}, L., {Schaeffer}, H.: {On the Convergence of the {SINDy} Algorithm}.
  arXiv e-prints arXiv:1805.06445 (May 2018)

\end{thebibliography}

\end{document}